**Key Points:**
- A generative model is developed to capture spatial dependence in heat extremes and simulate unseen but statistically plausible events
- Unseen heat extremes may disproportionately affect socioeconomically unprepared countries across the broader Middle East and North Africa
- Differentiating direct-hit and near-miss extremes reveals spatially structured, shifting hazard patterns that call for forward-looking adaptation planning

**Supporting Information:**

Supporting Information may be found in the online version of this article.

**Correspondence to:**
X. He and D. Zhang,
hexg@nus.edu.sg;
zhangdx@sustech.edu.cn

**Citation:**
Liu, X., Peng, X., Yan, S., Chen, Y., Zhang, D., Niu, Z., et al. (2026). Capturing unseen spatial heat extremes through dependence-aware generative modeling. *Earth's Future*, *14*, e2026EF008861. https://doi.org/10.1029/2026EF008861

Received 21 APR 2026
Accepted 17 AUG 2026

**Author Contributions:**
**Conceptualization:** Xiaogang He
**Data curation:** Xinyue Liu
**Formal analysis:** Xinyue Liu
**Funding acquisition:** Xiaogang He
**Investigation:** Xinyue Liu, Xiao Peng, Xiaogang He
**Methodology:** Xinyue Liu, Xiao Peng, Xiaogang He

# Capturing Unseen Spatial Heat Extremes Through Dependence-Aware Generative Modeling

**Xinyue Liu[1,2], Xiao Peng[1], Shuyue Yan[1], Yuntian Chen[3], Dongxiao Zhang[2,3], Zhixiao Niu[1], Hui-Min Wang[1], and Xiaogang He[1]**

[1]Department of Civil and Environmental Engineering, National University of Singapore, Singapore, Singapore, [2]School of Environmental Science and Engineering, Southern University of Science and Technology, Shenzhen, China, [3]Ningbo Institute of Digital Twin, Eastern Institute of Technology, Ningbo, China

**Abstract** Observed records of climate extremes provide an incomplete view of plausible hazards, missing "unseen" events beyond historical experience. Ignoring spatial dependence further underestimates hazards striking multiple locations simultaneously. We introduce DeepX-GAN (Dependence-Enhanced Embedding for Physical eXtremes—Generative Adversarial Network), a deep generative model that explicitly captures the spatial structure of rare extremes. Its zero-shot generalizability enables the simulation of statistically plausible extremes beyond the observed record, evaluated against long climate model large-ensemble simulations. We define two unseen types: direct-hit extremes that affect the target, and near-miss extremes that narrowly miss. These unrealized events reveal hidden risks and can either prompt proactive adaptation or reinforce a false sense of resilience. Applying DeepX-GAN to the Middle East and North Africa shows that the probability of unseen heat extremes is disproportionately distributed toward countries with high vulnerability and low socioeconomic readiness. Using a representative climate simulation, we demonstrate how future warming could expand and shift these hazards, creating persistent hotspots in Northwest Africa and the Arabian Peninsula and new hotspots in Central Africa, necessitating spatially adaptive resilience planning.

**Plain Language Summary** People typically learn about the dangers of climate extremes from past records. Yet these records can miss plausible but unobserved "unseen" extremes, leading to underestimation of risk, especially when spatially co-occurring events are ignored. We introduce a generative model that incorporates the spatial dependence structure of extreme events to simulate statistically plausible unseen extremes. We consider both unseen direct-hit (record-breaking events that strike a specific, highly exposed location) and near-miss (equally severe events that just happen to occur nearby) events. Direct hits show that "design for the past" is not enough. Near misses may mislead communities: if a severe event passes just to the side, people may feel previous warnings were "overreaction" and become less willing to act next time, even though a small shift in weather patterns could bring disaster directly over them. Applying the framework in the Middle East and North Africa under historical and future (SSP126, SSP585) climates using a representative climate model, we reveal substantial increases and spatial redistributions of unseen extremes. These hazards fall disproportionately on highly vulnerable, low-emitting countries, highlighting the need for spatially adaptive, forward-looking climate risk assessment and equitable adaptation planning.

## 1. Introduction

Recent record-shattering climate extremes have exposed a critical limitation in risk assessment: the overreliance on historical observations to define what is possible (Donat et al., 2020; Thompson et al., 2017). While extreme events are a natural part of a variable climate system (Coles, 2001), the short span of instrumental records in many regions means that plausible, high-impact scenarios remain unobserved and unaccounted for in infrastructure design and adaptation planning (Fischer et al., 2021; Kelder et al., 2022). This gap is particularly consequential in areas that, by chance, have not yet experienced the most severe manifestations of hazards (e.g., heatwaves), which we refer to as unseen extremes.

Unseen extremes are not speculative. Although absent from historical experience, they remain plausible within the underlying physical system and may be viewed as gray swans (Figure 1a) (De Marzo et al., 2022). Growing evidence demonstrates that extreme events previously deemed unlikely are indeed possible (Fischer et al., 2023; Gessner et al., 2021; Kelder et al., 2022; Peng et al., 2025; Thompson et al., 2017), or at minimum, their occurrence cannot be ruled out. Events like the 2021 Pacific Northwest heatwave, which shattered previous

**Project administration:** Xiaogang He
**Supervision:** Xiaogang He
**Validation:** Xinyue Liu
**Visualization:** Xinyue Liu, Shuyue Yan, Xiaogang He
**Writing – original draft:** Xinyue Liu, Shuyue Yan
**Writing – review & editing:** Xinyue Liu, Xiao Peng, Shuyue Yan, Yuntian Chen, Dongxiao Zhang, Zhixiao Niu, Hui-Min Wang, Xiaogang He

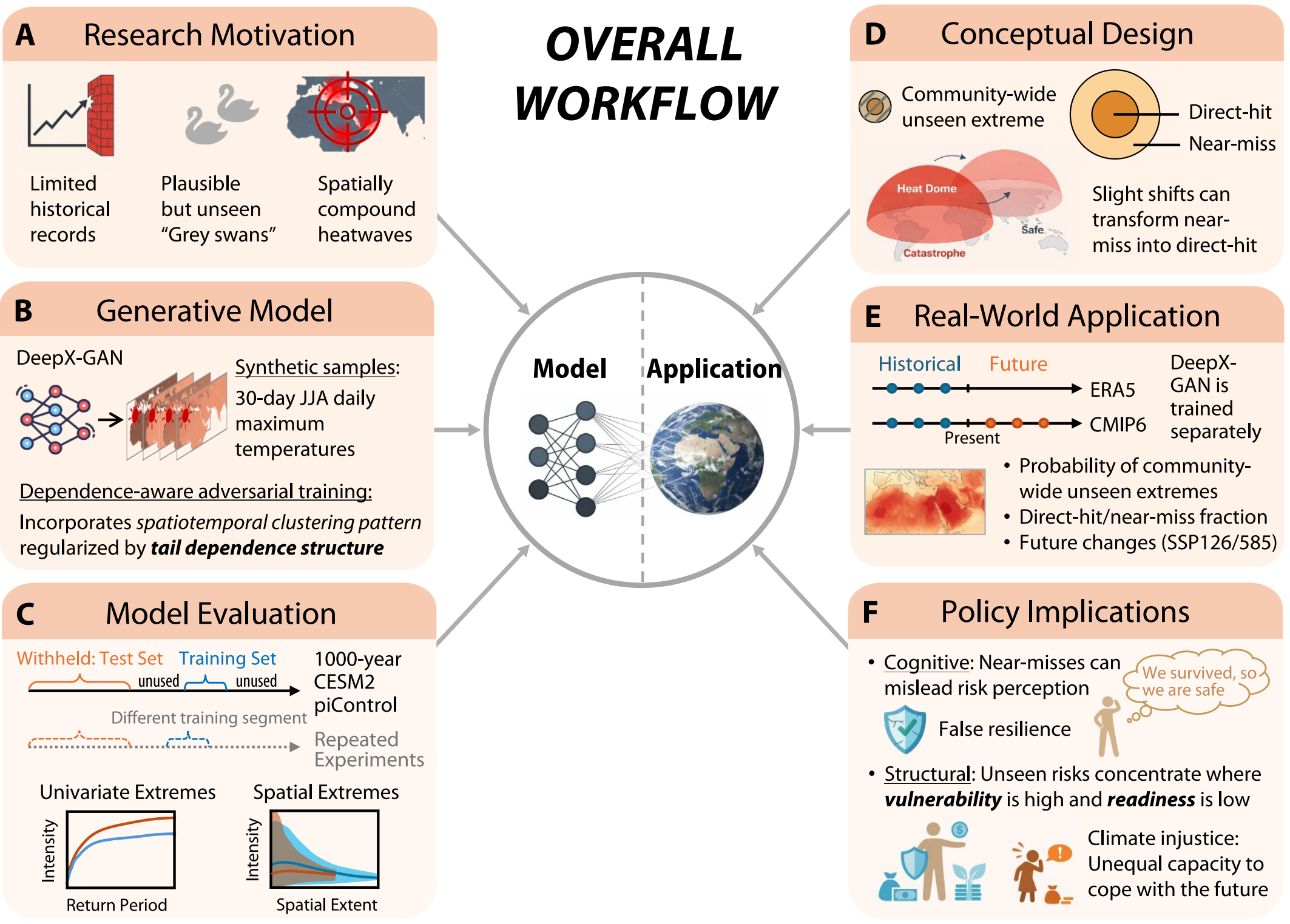

**Figure 1.** Overall workflow of the DeepX-GAN framework for unseen extreme analysis. The model components (a, b, c) establish a dependence-aware generative framework designed to represent spatially compounding extremes beyond the limits of short observational records, and rigorously evaluate its reliability using long (i.e., 1,000 years) preindustrial control simulations as the full climate distribution to test whether generated unseen extremes are statistically indistinguishable. The application components (d, e, f) deploy the model to historical reanalysis (ERA5) and future climate simulations (CMIP6) to estimate community-wide unseen extreme probabilities and direct-hit versus near-miss fractions, enabling interpretation of how these unprecedented hazards relate to national vulnerability and readiness regimes. See explanation for the acronyms in the Methods and Table S1 in Supporting Information S1.

records by up to 5°C and exceeded earlier statistical upper bounds derived from limited records, illustrate how such extremes can occur unexpectedly (Thompson et al., 2023; White et al., 2023). Their absence from past records can delay adaptation efforts, foster overconfidence in perceived resilience, and leave vulnerable populations at risk (Fouillet et al., 2008; Thompson et al., 2023).

We focus on two types of plausible but unrecorded events: (a) unseen direct-hit extremes (Figure 2a), which exceed historical values at high-exposure locations of interest, and (b) unseen near-miss extremes (Figure 2b), which narrowly bypass such locations but occur in adjacent areas. While near-miss extremes do not directly impact the target location as direct-hit extremes, we cannot rule out the possibility that future events might cause damages in the target location because of the spatial randomness, where small shifts in phenomena like heat domes—due to the stochastic nature of physical processes such as circulation changes (Brönnimann et al., 2025; Cai et al., 2024; Wicker et al., 2024) and land-atmosphere feedbacks (Duan et al., 2025)—can significantly alter the location and intensity of extreme events. Near-miss extremes can also trigger indirect, cascading socioeconomic disruptions across spatial boundaries. For instance, an extreme heatwave in an adjacent region can overload local healthcare systems, while localized infrastructure failures can rapidly ripple across the broader regional transportation network.

While direct hits reveal latent fragility, near misses may also shape risk perception and preparedness, either prompting proactive adaptation or reinforcing complacency, depending on how they are interpreted (Arvai et al., 2006; Dillon & Tinsley, 2008, 2016; Retchless & Ashley, 2022). When Hurricane Wilma was approaching the Florida Keys in 2005, fewer than 10% of residents followed evacuation instructions, influenced by prior "unnecessary" evacuations (Dillon et al., 2014). Yet this hurricane turned out to bring the highest storm surge since 1965, causing extensive property damage and community flooding (Pasch et al., 2006). For effective policy and decision-making, it is crucial to explore potential trajectories and assess the probabilities of both types of unseen extremes that either hit directly or are near misses.

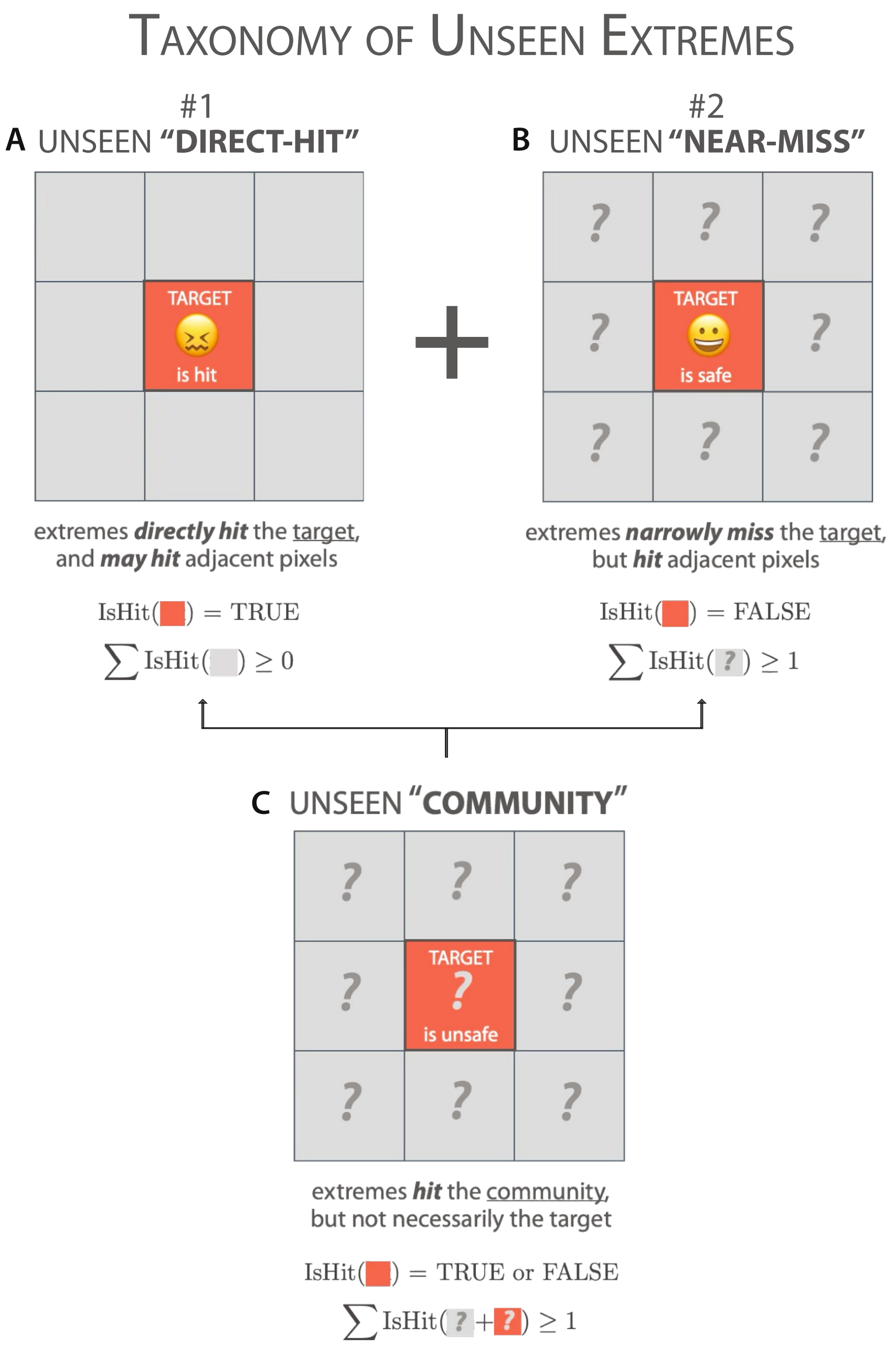

**Figure 2.** Schematic diagram for unseen direct-hit (a), near-miss (b), and community (c) extremes. Unseen extremes indicate events not observed historically but possible under climate stochasticity, which include direct-hit and near-miss extremes. Direct-hit (a) stands for extremes directly hitting the target location, and near-miss (b) indicates events avoided by chance despite occurring in surrounding regions. The probabilities of the two types of unseen extremes add up to the community-wide unseen extremes (c). IsHit() denotes the event outcome of unseen extremes occurring in a location. IsHit() = 1 if unseen extremes occur and IsHit() = 0 otherwise. In our analysis section, we compute the occurrence of unseen direct-hit (a) and near-miss (b) extremes relative to that of community-wide unseen extremes (c), so the direct-hit and near-miss fractions sum to one (see details in Methods). Note that the $3 \times 3$ neighborhood is shown solely for conceptual illustration, and the event classification scheme can accommodate flexible spatial scales. DeepX-GAN models spatial dependence across the entire study domain, unrestricted by this local structure.

Current approaches to modeling unseen extremes face significant challenges, especially the overlooked role of extreme events' spatial dependence structure. Spatially compounding events introduce complexity (Jiang et al., 2026; Wang & He, 2025), as their rarity confines them to a narrow corner of the multi-dimensional probability space and necessitates extensive data for robust assessment (Coles, 2001; B. Long et al., 2023). The lack of abundant observational data for these events impedes accurate estimation of their likelihood and subsequent socioeconomic impacts. Parametric statistical methods, such as extreme value analysis (Coles, 2001) and copula-based models (He et al., 2020; Sadegh et al., 2017; Salvadori et al., 2007), rely on assumptions and are often fitted to relatively short observational records. As a result, estimated upper-tail bounds may later be exceeded by new observations due to sampling variability (Thompson et al., 2023; L. Zhang et al., 2024; Thompson et al., 2022). Moreover, these approaches primarily characterize marginal extremes at individual locations and do not explicitly represent the spatial extent or organization of compound extreme events. Hybrid physics-based data-driven models (such as Single Model Initial-condition Large Ensemble [SMILE] (Bevacqua et al., 2023; Lehner, 2024; Maher et al., 2021), ensemble boosting (Fischer et al., 2023; Gessner et al., 2021, 2023), and UNprecedented Simulated Extremes using ENsembles [UNSEEN] (Kelder et al., 2020, 2022; Thompson et al., 2017, 2019)) offer alternatives by introducing randomness to simulate climate extremes in large ensembles (Aalbers et al., 2018; Leduc et al., 2019; Rampal et al., 2025; von Trentini et al., 2019), yet these methods require meticulous model setup (Fischer et al., 2023; Gessner et al., 2021; Kelder et al., 2020) or impose high computational demand (Bevacqua et al., 2023; Kelder et al., 2020, 2022; Maher et al., 2021; Thompson et al., 2017, 2019).

Recent advances in Artificial Intelligence (AI) offer tremendous opportunities (e.g., Kochkov et al., 2024; Price et al., 2025; Watt-Meyer et al., 2025) for computationally efficient simulation of spatially compounding extremes. The challenge of limited data, known as the "data wall" in the AI industry, can be jumped over by machine-generated synthetic data sets (The Economist, 2024). Previous efforts addressing small sample size for extreme events have included tailoring loss functions to emphasize outliers (Ding et al., 2019; M. Zhang et al., 2021; Hess & Boers, 2022) and transforming skewed distributions for improved learning (Boulaguiem et al., 2022; Huster et al., 2021). However, these methods often overlook the crucial spatial dependencies inherent in synchronized events. Moreover, these methods typically generate climate realizations as either one-dimensional time series (Ding et al., 2019; M. Zhang et al., 2021) or static two-dimensional spatial fields (Bhatia et al., 2021; Boulaguiem et al., 2022; Goodfellow et al., 2014), neglecting temporal evolution patterns in spatial fields that characterize climate dynamics in three-dimensional space.

Here we present DeepX-GAN (Dependence-Enhanced Embedding for Physical eXtremes—Generative Adversarial Network), a deep generative framework that integrates spatial extremal dependence to better capture spatially compounding and unseen extremes (Figure 1). This framework builds on Generative Adversarial Network (GAN, Figure S1 in Supporting Information S1), for its ability to model high-dimensional data distributions via dynamic adversarial training without requiring pre-defined likelihood functions (Bishop & Bishop, 2024; Goodfellow et al., 2014; Tomczak, 2022). However, traditional GANs are unable to perform well on spatially compounding extremes, as these rare samples can be diluted by less extreme samples during training. To overcome this, we propose to incorporate spatial extremal dependence structure through an embedding loss based on the DeepX metric (Dependence-Enhanced Embedding for Physical eXtremes), which detects how local anomalies align across space, with greater emphasis placed on locations that tend to experience extremes together. This constraint allows the model to learn how extreme values co-occur across space, and the model turns out to generalize better for out-of-distribution samples. We assess DeepX-GAN's generalizability to unseen extremes through experiments inspired by zero-shot learning (Larochelle et al., 2008), where models are evaluated to generate extremes not encountered during training. This mimics real-world scenarios where extreme conditions could differ in magnitude and spatiotemporal patterns from historical observations.

We demonstrate DeepX-GAN's ability to generate unseen yet statistically plausible and physically realistic heat extremes in the Middle East and North Africa (MENA), a region identified as the global hotspot of human vulnerability to climate change. According to climate vulnerability indices (University of Notre Dame, 2023), the 10 countries ranked as the most susceptible to climate change are all located within the extensive MENA region. These communities endure amplified consequences from spatially compounding extremes due to constrained local resources and high reliance on international support (Wang et al., 2024). Moreover, the lack of high-quality observational data in MENA hinders assessment and preparedness for climate extremes due to underdeveloped monitoring infrastructure. Therefore, the MENA region serves as an ideal testbed for extreme data augmentation and risk assessment.

We apply DeepX-GAN to simulate and augment heat extremes across the extensive MENA region and compare historical (1979–2014) and future (2065–2100) periods in terms of community-wide unseen extreme probability, direct-hit fraction, and near-miss fraction under both high-emission (SSP585) and mitigated (SSP126) scenarios. The spatial distribution of these unseen extremes highlights a profound inequity: less developed, highly vulnerable countries are likely to face a disproportionately high chance of occurrence despite minimal contributions to global emissions, potentially exacerbating climate injustice. Using a representative climate simulation, we find that future climate change may substantially elevate and redistribute unseen extreme probabilities, particularly over central Africa, even under the mitigated scenario. This distribution shift underscores the need to differentiate direct-hit and near-miss extremes to develop spatially adaptive policies that anticipate emergent hotspots rather than simply extrapolating from historical patterns.

## 2. Materials and Methods

### 2.1. Overall Framework

We develop a dependence-aware deep generative framework to enhance the simulation of spatially compounding climate extremes while augmenting spatiotemporal climate data sets (Figure 1). Accurately representing these compounding events requires capturing extremal dependence across space. DeepX-GAN addresses this by explicitly incorporating pixel-level spatial extremal dependence through an embedding metric. Because extremes are rare in time but occur across space, learning their spatial co-occurrence may allow the model to “borrow” information from other locations. This mechanism shares a similar intuition with the concept of space-for-time substitution, where spatial variation is used to gain insight into processes that are undersampled temporally (Lovell et al., 2023). By incorporating this spatial extremal dependence as a constraint during training, the model gains additional information about tail behavior, making it possible for the generation of unprecedented extremes at a given location while preserving realistic spatial structure. It should be noted that although we use a GAN architecture, the proposed method is model-agnostic and can be readily integrated into other generative frameworks.

We demonstrate DeepX-GAN's superior performance compared to the baseline model after explicitly accounting for the spatial dependence structure of climate extremes in DeepX-GAN's model architecture. We also design “unseen” experiments in a zero-shot learning context using long stationary climate simulations to assess DeepX-GAN's ability to represent plausible but historically unobserved extreme events. In the application, we specifically examine two types of unseen extremes: unseen direct-hit extremes and unseen near-miss extremes, assessing their probabilities for historical period and two future scenarios. It should be noted that our analysis focuses primarily on the hazard dimension of risk while incorporating socioeconomic factors for impact-related discussions, whereas a full risk assessment would require additional consideration of vulnerability and exposure.

### 2.2. Data Sets

#### 2.2.1. Climate Reanalysis

To investigate heat extremes, we select the daily maximum 2-m air temperature ($T_X$) as the primary variable. Historical period spans from 1979 to 2014, covering the extensive MENA region (22°W–98°E and 7°S–53°N), which includes the world's 10 most vulnerable countries to the adverse impacts of climate change (IPCC, 2022; University of Notre Dame, 2023). The temperature data are obtained from the latest European Centre for Medium-Range Weather Forecasts (ECMWF) Reanalysis v5 (ERA5) data set. This reanalysis product has consistent, long-term data records and is widely used by the climate research community. We train our model using 2,268 sequences constructed by applying a 30-day sliding window (with a one-day stride) to the 36 boreal summers (June–July–August, JJA) during 1979–2014. The spatial resolution is approximately $1.9^\circ \times 1.9^\circ$.

#### 2.2.2. Preindustrial Model Simulations

To evaluate the reliability of generated heat extremes, we use $T_X$ from 1,000-year preindustrial climate model simulations of the Community Earth System Model version 2 (CESM2) Large Ensemble. These long, unforced simulations provide a reference for a stationary full climate distribution against which unseen extremes can be assessed. Detailed experimental design is documented in the following "Unseen Extremes" section.

#### 2.2.3. Climate Model Projections

For future scenarios under climate change, we utilize $T_X$ from the Coupled Model Intercomparison Project Phase 6 (CMIP6) simulations. Specifically, we select CMCC-ESM2 (the second-generation Earth System Model developed at the Euro-Mediterranean Centre on Climate Change) as a representative model to estimate future changes in unseen extremes, because it well represents the dynamic atmosphere-ocean-land interactions (Cherchi et al., 2019; Lovato et al., 2022) and simulate a consistent increase in global mean temperature over 1975–2014 with HadCRUT5 (Lovato et al., 2022), the new version of the Met Office Hadley Centre Climatic Research Unit global surface temperature data set.

We train our model on historical simulation (1979–2014) and future projections (2065–2100) under two greenhouse gas emissions pathways (SSP126, optimistic development; SSP585, fossil-fuel-based development), in a manner similar to that for reanalysis data.

#### 2.2.4. Vulnerability and Readiness Indicators

Country-level vulnerability and readiness indicators are obtained from Notre Dame Global Adaptation Initiative's (ND-GAIN) Country Index to assess where the greatest needs and opportunities exist for improving resilience to climate change (University of Notre Dame, 2023). The vulnerability indicators are considered from six sectors: food, water, health, ecosystem, habitat, and infrastructure. The readiness indicators encompass three dimensions: economic readiness, governance readiness, and social readiness. ND-GAIN scales these indicators from 0 to 1, where 0 represents the globally lowest computed value, and 1 indicates the highest.

### 2.3. Deep Generative Model

DeepX-GAN explicitly incorporates extremal dependence structures in climate fields. We choose the Generative Adversarial Network (GAN) as the backbone for the deep generative framework because GAN does not rely on prescribed distribution groups or estimates of likelihood, unlike other deep generative models such as Variational Auto-Encoders (Tomczak, 2022). Instead, GANs utilize implicit density models represented by neural networks and can tackle intractable high-dimensional probabilistic distributions (Bishop & Bishop, 2024; Goodfellow et al., 2014; Tomczak, 2022). Our DeepX-GAN follows the traditional GAN architecture with two components: a generator that learns to reproduce the original distribution from random noise, and a discriminator that learns to distinguish between target and generated distributions (Figure S1 in Supporting Information S1). However, our model uniquely incorporates an additional input channel for an embedding metric that accounts for both spatiotemporal autocorrelation and extremal correlation, capturing complex interactions across space and time (see details below).

Throughout this study, the term DeepX-GAN refers to our proposed deep generative model framework in general. To ensure clarity when discussing different phases of our evaluation and application, we use specific naming conventions for the model variants based on their training data: DeepX-GAN-ERA5, DeepX-GAN-CESM2, and DeepX-GAN-CMCC-ESM2 denote the models trained on the ERA5 reanalysis, the CESM2 Large Ensemble, and the CMCC-ESM2 data sets, respectively.

#### 2.3.1. Baseline Model

DeepX-GAN is grounded in the architecture of SPATE-GAN (SPAtioTEmporal association Generative Adversarial Network) (Klemmer et al., 2022), which is designed to generate realistic spatiotemporal simulations, such as weather patterns or traffic flow. SPATE-GAN utilizes the principles of causal optimal transport, a method that quantifies the transformation of one set of data points into another over time, while considering the inherent cause-and-effect relationships. This approach is integrated into the adversarial framework of the model through the development of mixed Sinkhorn loss, which guides the generator to learn the real distribution in the most cost-effective way. The model also incorporates a unique autoregressive embedding metric, named SPATE (SPAtioTEmporal association), to detect the spatiotemporal clustering patterns and steer the model toward learning spatiotemporal patterns. Empirical evidence has demonstrated that SPATE-GAN outperforms the previous models in capturing the spatiotemporal dynamics (Klemmer et al., 2022).

### 2.3.2. Dependence-Aware Deep Generative Model

We enhance the baseline model by infusing the knowledge of the spatial dependence structure of extreme events into the adversarial training process. This is achieved by embedding a novel DeepX metric, which identifies evolving patterns across space and time while explicitly accounting for the spatial correlation of extreme events. By feeding DeepX as an extra channel in the discriminator (Figure S1 in Supporting Information S1), discrepancies in spatial extremal dependence directly contribute to the adversarial objective. This integration forces the generator to align its synthetic samples with real-world inputs within a transformed space where complex, non-linear spatiotemporal patterns of extreme events are easier to learn. Consequently, DeepX-GAN gains insight into the collective behavior of spatial extremes, facilitating more reliable simulation of spatially compounding events.

Mathematically, our DeepX metric tracks spatiotemporal autocorrelation by measuring how observations deviate from their expected values in both space and time (i.e., space-time expectation), then assessing these deviations against nearby observations to identify areas of notable change or homogeneity. Meanwhile, DeepX assigns greater weights to locations that are linked during extreme events to account for spatial extremal dependence structure. For pixel $i$ and time step $t$, the $\mathrm{DeepX}_{it}$ metric is defined as:

$$\mathrm{DeepX}_{it} = \frac{(n-1)z_{it}}{\sum_{j=1}^{n} z_{jt}^2} \sum_{j=1, j \neq i}^{n} w_{ij} z_{jt} \tag{1}$$

where $n$ is the number of pixels per time snapshot, and $z_{it}$, $z_{jt}$ are deviations of values $x_{it}$, $x_{jt}$ at pixels $i$, $j$ and time step $t$ from their expected values $\mu_{it}^{\mathrm{mixed}}$, $\mu_{jt}^{\mathrm{mixed}}$, defined as:

$$z_{it} = x_{it} - \mu_{it}^{\mathrm{mixed}} \tag{2}$$

where $x_{it}$ is the value at pixel $i$ and time step $t$, $w_{ij}$ is a binary number indicating the spatial proximity of observations $x_{it}$ and $x_{jt}$, which is defined as:

$$w_{ij} = \begin{cases} 1 & \text{if } x_{jt} \text{ is in the neighborhood of } x_{it}, t \in \mathbb{Z}^+ \\ 0 & \text{elsewhere} \end{cases} \tag{3}$$

where $\mathbb{Z}^+$ represents the set of all positive integers.

The space-time expectation $\mu_{it}^{\mathrm{mixed}}$ is computed using spatial observations at the current time step $t$ and temporal observations at the current location $i$ in past time steps $t' < t$. It has two components regulated by the hyperparameters $\theta_{\mathrm{A}}$ and $\theta_{\mathrm{B}}$, that is,

$$\mu_{it}^{\mathrm{mixed}} = \theta_{\mathrm{A}} \mu_{it}^{\mathrm{A}} + \theta_{\mathrm{B}} \mu_{it}^{\mathrm{B}} \tag{4}$$

Note that when $\theta_{\mathrm{A}} = 1$ and $\theta_{\mathrm{B}} = 0$, the model reduces to the baseline SPATE-GAN, which only considers spatiotemporal autocorrelation without focusing on extremal dependence structure.

The space-time expectation $\mu_{it}^{\mathrm{A}}$ considers spatiotemporal coupling patterns (Figure S2 in Supporting Information S1), which is adapted from Klemmer et al. (2022):

$$\mu_{it}^{\mathrm{A}} = \frac{\sum_{j=1}^{n} x_{jt} \sum_{t' < t} b_{tt'} x_{it'}}{\sum_{j=1}^{n} \sum_{t' < t} b_{tt'} x_{jt'}} \tag{5}$$

Here, $b_{tt'} = \exp\left(-\frac{|t-t'|}{l}\right)$ is a weight adjusting the influence of past temporal information according to the time lag $|t - t'|$, and $l$ is the length scale of the exponential kernel. The modified space-time expectation $\mu_{it}^{\mathrm{B}}$ additionally incorporates spatial extremal dependence structure.

$$\mu_{it}^{\mathrm{B}} = \frac{\sum_{j=1}^{n} k_{ij,t}\chi_{ij}x_{jt}\sum_{t'<t} b_{tt'}x_{it'}}{\sum_{j=1}^{n}\sum_{t'<t} b_{tt'}x_{jt'}} \tag{6}$$

The coefficient $k_{ij,t}\chi_{ij}$ serves as a weight that modulates the impact of spatial information from pixel $j$ on pixel $i$. This modulation is based on the presence of extreme values at both pixels $(k_{ij,t})$ and the degree to which these extremes at the two locations exhibit correlation $(\chi_{ij})$.

$k_{ij,t}$ is formulated as:

$$k_{ij,t} = \begin{cases} 1 & \text{if } F_i(x_{it}) > q \text{ and } F_j(x_{jt}) > q \\ 0 & \text{elsewise} \end{cases} \tag{7}$$

where $F_i$ and $F_j$ are cumulative distribution functions (CDFs) for data at locations $i$ and $j$, respectively; $q$ is a predefined extreme threshold in the range of 0–1.

The extremal correlation, or upper tail dependence coefficient used in Equation 6, is formally defined for locations $i$ and $j$ as:

$$\chi_{ij} = \lim_{q\to 1} P(F_i(x_i) > q | F_j(x_j) > q) \in [0,1] \tag{8}$$

When $\chi_{ij} = 0$, locations $i$ and $j$ are asymptotically independent, indicating no tendency for concurrent extreme events. Extremal correlation could be interpreted as the probability of observing an extreme event at one location, given that an extreme event occurs at another location. Using conditional probability, we empirically compute the extremal correlation for pixels $i$ and $j$ as:

$$\chi_{ij} = \frac{P(F_i(x_i) > q, F_j(x_j) > q)}{P(F_j(x_j) > q)} \tag{9}$$

## 2.4. Unseen Extremes

### 2.4.1. Experimental Design

Long, unforced climate simulations allow direct evaluation of unseen extremes against a withheld reference distribution. We use the 1,000-year CESM2 Large Ensemble preindustrial control (piControl) simulation as a representation of the full unforced climate distribution and emulate partial observability by randomly sampling short consecutive segments (Figure S3 in Supporting Information S1). A fixed 660-year portion of the simulation is reserved exclusively for this evaluation and is not used in any training experiment. For each experiment, a 36-year segment—comparable in length to the ERA5 record—is randomly selected from the remaining portion of the simulation for training. DeepX-GAN trained on the short segment is then assessed on its ability to generate extremes beyond the training record while remaining statistically similar to the 660-year withheld distribution. This procedure is repeated six times using different training segments to assess robustness to sampling variability.

### 2.4.2. Unseen "Direct-Hit" and "Near-Miss" Extremes

Our work differentiates between two types of unseen extremes. An unseen "direct-hit" extreme (Figure 2a) occurs when an unprecedented climate event directly hits a location of interest (e.g., a capital city). On the contrary, an unseen "near-miss" extreme (Figure 2b) occurs when such an event narrowly misses the target location but comes close enough to it.

For quantitative assessment, we distinguish between the probability of community-wide unseen extremes (Figure 2c) and the fractions of direct-hit and near-miss events within those unseen extremes. For each location in a community, we define the record-breaking threshold based on a consistent severity level, computed as the local return level corresponding to the return period of the highest temperature observed in the training record at the target location.

The community-wide unseen extreme probability, $P^{\text{community}}$, is defined as the probability that an extreme event at least as severe as the record-breaking threshold occurs within the community:

$$P^{\text{community}} = P(\exists_{i \in \{\text{neighbors}\} \cup \{\text{target}\}} I_i \geq \alpha_i) \tag{10}$$

where $i$ is the index of locations, belonging to ($\in$) the union ($\cup$) of the neighbor set ({neighbors}) and the target location ({target}). $I_i$ denotes the event intensity at location $i$, and $\alpha_i$ is the exceedance threshold of intensity at that location. The symbol $\exists$ expresses the logical existence of a condition, meaning that the statement is satisfied if at least one location in the community fulfills $I_i \geq \alpha_i$.

Conditioned on the occurrence of community-wide unseen extremes, we further partition events into direct-hit and near-miss outcomes. A direct-hit unseen event occurs when the target location itself exceeds the record-breaking threshold. The corresponding fraction, $R^{\text{direct-hit}}$, is defined as

$$R^{\text{direct-hit}} = \frac{P(I_{\text{target}} \geq \alpha_{\text{target}})}{P^{\text{community}}} \tag{11}$$

where $I_{\text{target}}$ is the event intensity at the target location, and $\alpha_{\text{target}}$ is the corresponding exceedance threshold of intensity at that location.

A near-miss unseen event occurs when the target location does not exceed its threshold, but at least one neighboring location does. The corresponding fraction, $R^{\text{near-miss}}$, is given by

$$R^{\text{near-miss}} = \frac{P(\exists_{i \in \{\text{neighbors}\}} I_i \geq \alpha_i \wedge I_{\text{target}} < \alpha_{\text{target}})}{P^{\text{community}}} \tag{12}$$

where $\wedge$ denotes a logical "AND," that is, the statement is only true if and only if both of its operands are true.

By construction, the direct-hit and near-miss fractions sum to unity and describe the relative partitioning of community-wide unseen extremes. These quantities can be estimated empirically from generated data by DeepX-GAN or derived analytically under idealized assumptions.

For a null model that assumes spatial independence (i.e., spatial extremal correlation = 0), where heat extremes occur randomly and independently at each pixel with a fixed occurrence probability, based on the definition in Equations 10–12, the direct-hit and near-miss fractions are calculated as:

$$R^{\text{direct-hit}}_{\text{random}} = \frac{P_{\text{target}}}{1 - \prod_{i \in \{\text{neighbors}\}} (1 - P_i)(1 - P_{\text{target}})} \tag{13}$$

$$R^{\text{near-miss}}_{\text{random}} = \frac{\left[1 - \prod_{i \in \{\text{neighbors}\}} (1 - P_i)\right](1 - P_{\text{target}})}{1 - \prod_{i \in \{\text{neighbors}\}} (1 - P_i)(1 - P_{\text{target}})} \tag{14}$$

where $P_i$ is the exceedance probability for neighbor $i \in$ {neighbors}, and $P_{\text{target}}$ represents the exceedance probability for the target pixel. The product operator $\prod$ denotes multiplication over all neighboring locations.

Under an idealized stationary and exchangeable process, the probability of observing an event more severe than the historical maximum is expected to be $1/(S+1)$ (i.e., $P_i = P_{\text{target}} = 1/(S+1)$), where $S$ denotes the length of the known period. Note that because short records introduce sampling variability, this theoretical probability $1/(S+1) = 1/(36+1) \approx 0.027$ for 36-year data may not reflect the true exceedance probability (e.g., if the historical period captured unusually extreme or mild events), yet we use it to define the hotspots for community-wide unseen extremes. The direct-hit and near-miss fractions are then compared with a spatial independent null model, where the unseen direct-hit and near-miss fractions calculated by Equations 13 and 14 based on this theoretical exceedance probability are $R_{\text{random}}^{\text{direct-hit}} = 0.12$ and $R_{\text{random}}^{\text{near-miss}} = 0.88$, respectively. Here, we use a $3 \times 3$ grid of pixels to identify neighboring locations, but the definitions of unseen community, direct-hit, and near-miss events can also be applied to flexible spatial scales.

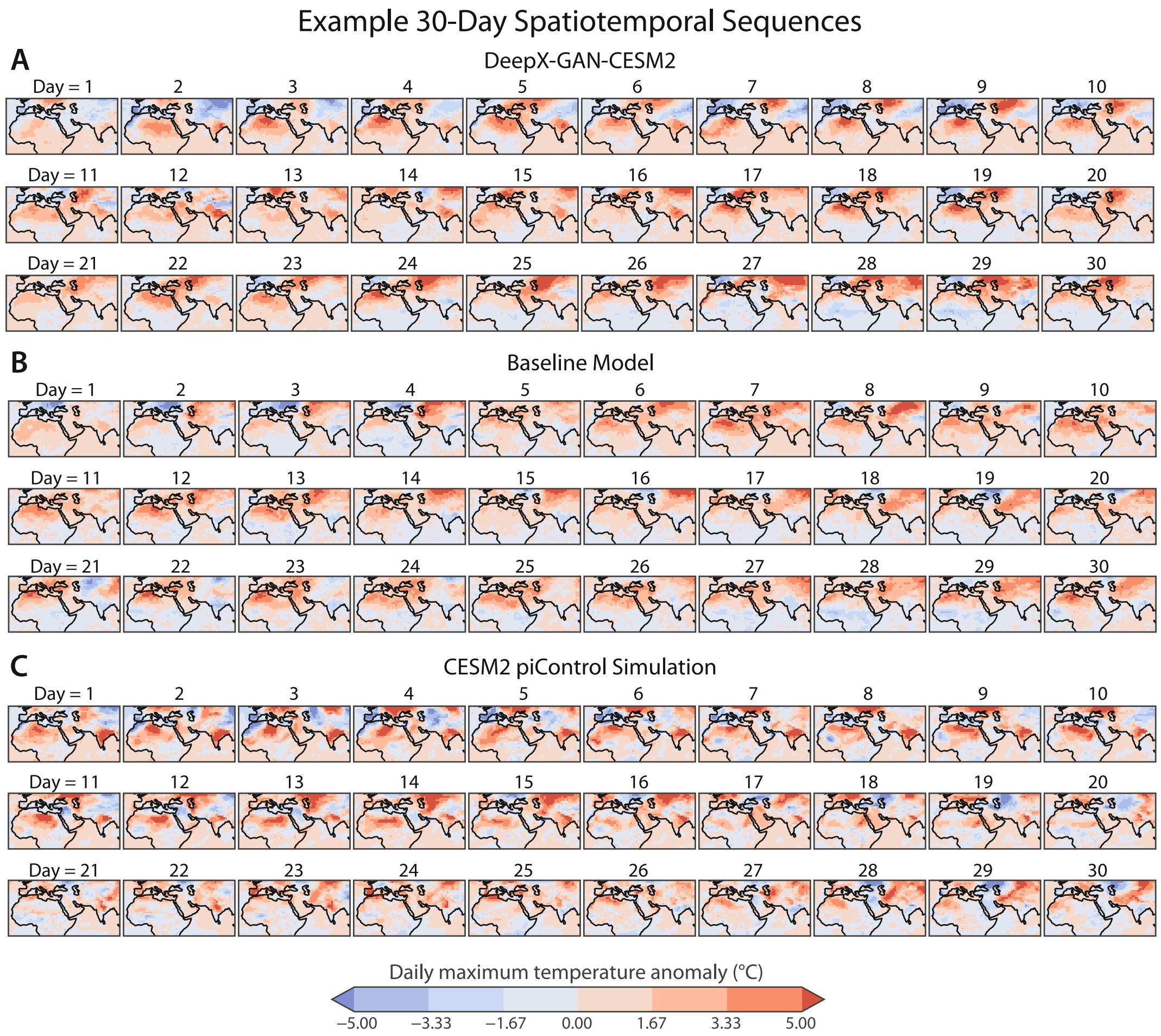

**Figure 3.** Example sequences of 30-day daily maximum temperature anomalies. Data are generated by DeepX-GAN-CESM2 (a), the baseline model (b), and from the training record of CESM2 piControl simulation (c). Note that these examples illustrate pattern-level behavior and are not meant to reproduce specific individual events.

For country-level analysis, we aggregate pixel-level unseen extreme metrics by averaging across each country. Our analysis includes 77 countries within the study region, with each selected country covering at least two pixels.

## 3. Results

We compare our proposed model with a baseline generative model that enforces spatiotemporal coherence but does not explicitly account for extremal dependence (Klemmer et al., 2022). Both models are trained on daily maximum temperature over 30-day summer windows in the MENA region, with the goal of simulating the overall temperature distribution and heat extremes, rather than reproducing specific events. We evaluate the reliability of generated unseen extremes and the performance relative to the baseline model using CESM2 simulations (Figures 3 and 4), which provide a long, physics-based reference distribution for extreme statistics, before using the generative framework on ERA5 data set (Figure 5) and CMCC-ESM2 simulations (Figure 6) to estimate unseen extreme probabilities in real-world applications (see Figure 1 for a complete workflow).

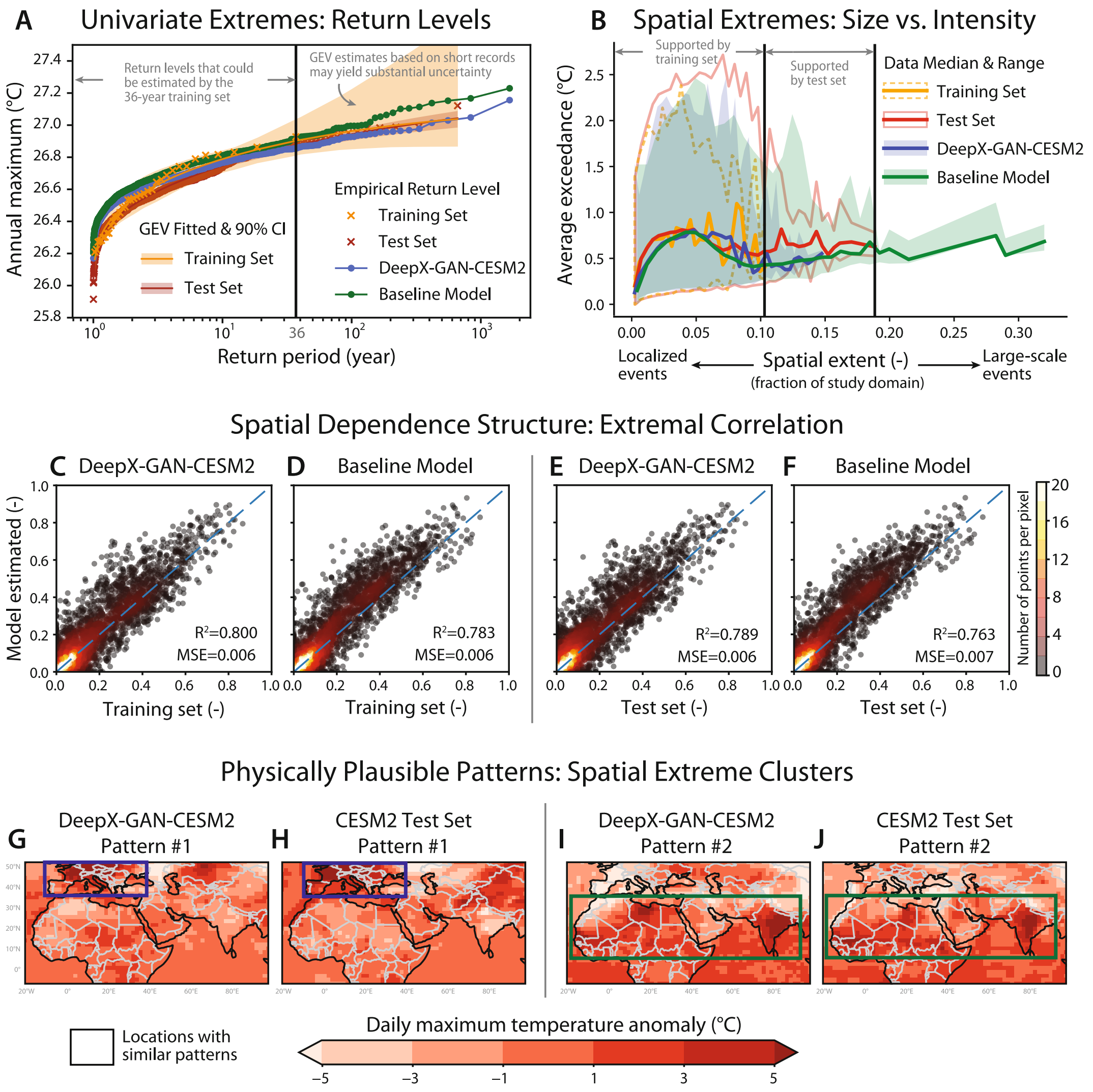

**Figure 4.** Evaluation of model performance in generating heat extremes beyond the training record. Univariate extreme behavior is evaluated using return levels of domain-averaged annual maximum temperature (a), where estimates from the 36-year training set, the withheld 660-year test set, DeepX-GAN-CESM2 generated samples, and the baseline model are compared. Spatial extreme characteristics are examined through the relationship between spatial extent and mean intensity above a high-temperature threshold (b; local 95th percentile), summarized by median and range ([0.5th, 99.5th]) across ensembles. The representation of spatial dependence is assessed using pairwise extremal correlations (c–f) between generated samples and the training and test data sets, for both DeepX-GAN-CESM2 and the baseline model. Examples of spatial extreme event clusters (g–j) compare generated samples (g, i) with test set patterns (h, j), with rectangles marking regions with similar large-scale patterns.

### 3.1. Generating Unseen yet Plausible Extremes

In this study, we focus on unseen extremes absent from the training data due to sampling variability arising from limited records. To evaluate the reliability of such unseen extremes, we design a physics-grounded validation framework using the CESM2 Large Ensemble piControl simulation (Figure S3 in Supporting Information S1), which provides 1,000-year unforced climate samples. We train DeepX-GAN-CESM2 on a randomly selected 36-year segment (matching the ERA5 reanalysis historical period length of 1979–2014) and evaluate its ability to generate extremes that extend beyond the training record while remaining statistically similar to an independent withheld reference distribution (so-called zero-shot setting). A fixed 660-year portion of the simulation is reserved exclusively for this evaluation and is not used in any training experiment. This procedure is repeated six times using different randomly selected 36-year segments to assess robustness to sampling variability.

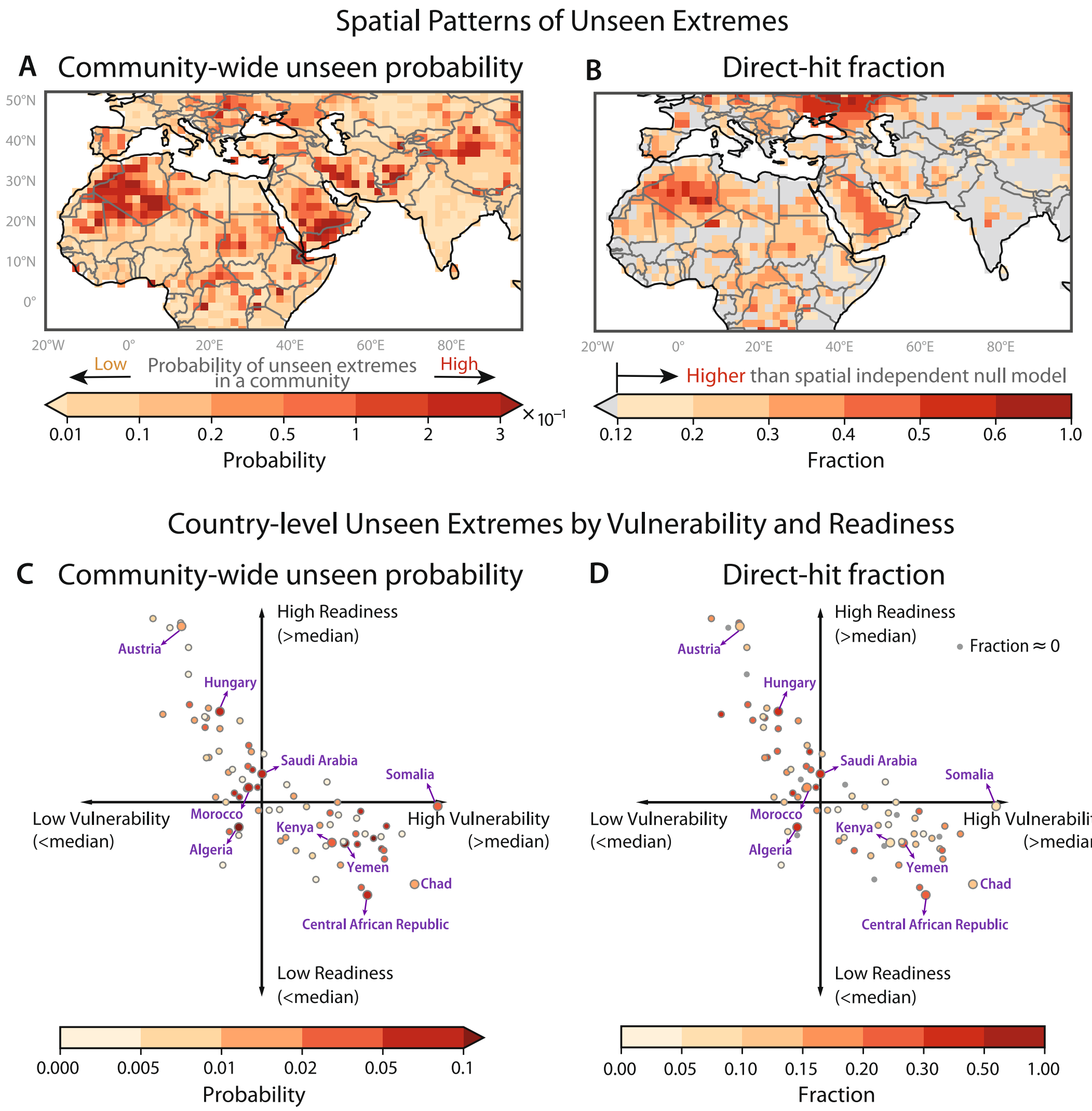

**Figure 5.** Model-estimated likelihood of extremes unseen in the historical (1979–2014) record. Pixelwise estimates of community-wide unseen extreme probability (a) and direct-hit fraction (b) are derived from DeepX-GAN-ERA5 trained on ERA5 reanalysis data (1979–2014), and aggregated to country-level statistics for community-wide probability (c) and direct-hit fraction (d), shown in relation to national vulnerability and readiness indicators. Gray shading in (b) indicates locations where the direct-hit fraction falls below that expected under a simple null model that assumes spatial independence and randomness. Gray points in (d) denote no estimated events of unseen direct-hit extremes. The countries discussed in the main sections are highlighted and named. The near-miss fraction is not shown because the direct-hit and near-miss fractions are complementary and sum to one at each target location, as illustrated in Figure 2.

### 3.1.1. Spatiotemporal Patterns

To illustrate that the generated samples reproduce realistic spatial and temporal patterns, we present one representative 30-day sequence for each model (Figures 3a–3c). These examples are shown for illustrative purposes; additional realizations exhibit similar qualitative behavior. Because the sequences are independent realizations rather than replications of specific CESM2 events, one-to-one correspondence is not implied, and comparisons focus on pattern-level behavior.

Both DeepX-GAN-CESM2 and the baseline model produce spatially coherent and temporally persistent temperature anomaly fields, indicating that generative sampling does not introduce obvious artifacts or unrealistic variability. For the sequence generated by DeepX-GAN-CESM2 (Figure 3a), elevated anomalies tend to organize into contiguous regional patches that persist across multiple consecutive days (e.g., days 4–10 and 22–30). In contrast, the baseline sequence (Figure 3b) exhibits comparatively more spatially diffuse patterns, with weaker gradients and less contrast within positive anomaly patches (e.g., days 1–7 and 24–30). Besides, both models display small deviations from CESM2 in the mean, standard deviation, and 95th percentile (Figure S4 in Supporting Information S1). We also assess the dispersion of the generated ensemble using a rank histogram (Hamill, 2001), which shows a near-uniform distribution (Figure S5 in Supporting Information S1). This demonstrates that DeepX-GAN-CESM2 maintains realistic spatiotemporal patterns and first-order behaviors. In the following sections, we will further evaluate the model's higher-order improvements in extreme structure.

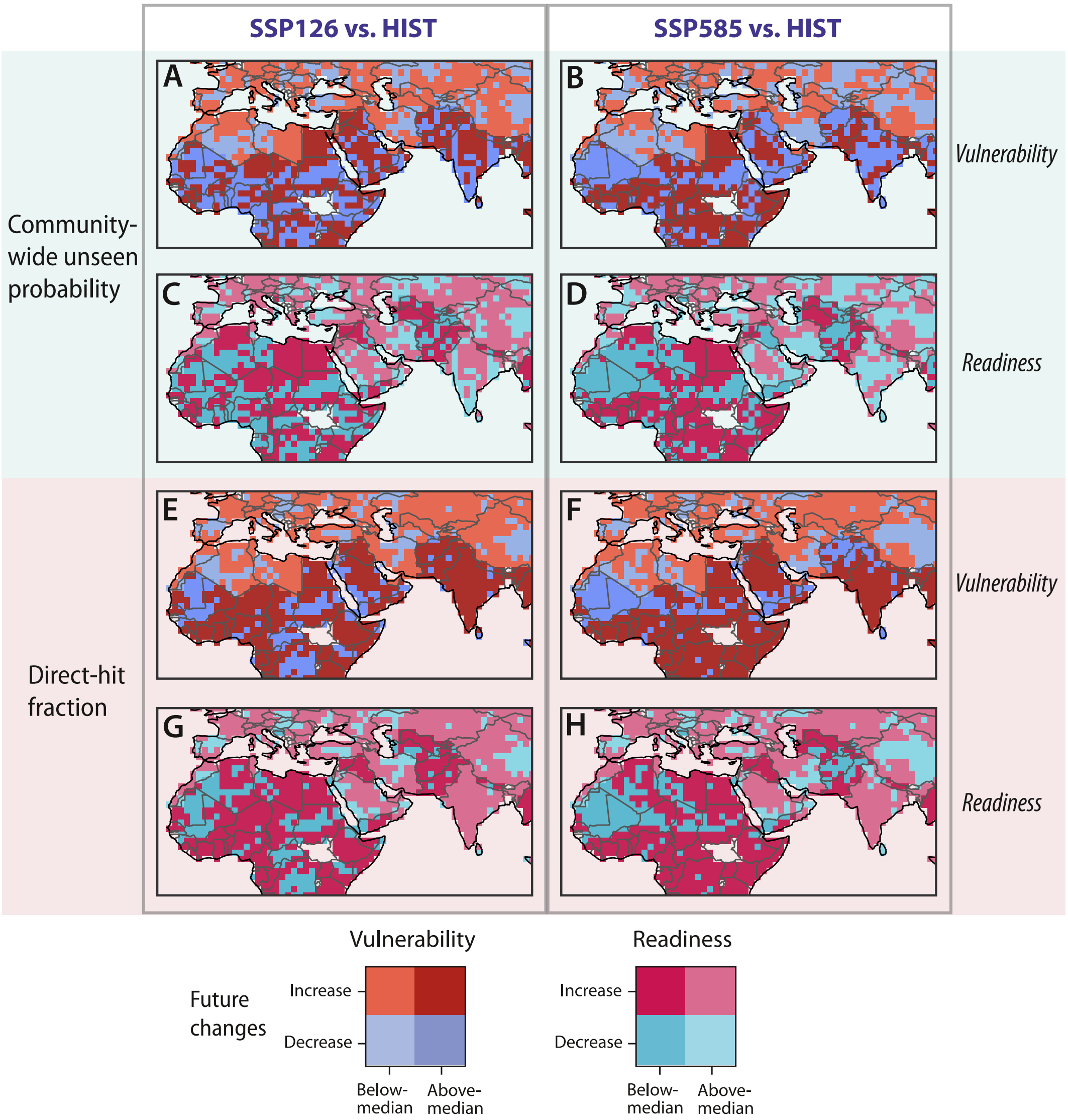

**Figure 6.** Future (2065–2100) pixelwise probability/fraction of unseen extremes overlaid on vulnerability and readiness indicators. The changes compared to the historical climate (1979–2014) in community-wide unseen extremes (a–d) and direct-hit extremes (e–h) under future SSP126 (a, c, e, g) and SSP585 (b, d, f, h) scenarios are estimated from ensembles generated by DeepX-GAN-CMCC-ESM2. We do not show near-miss extremes here because, in our definition, direct-hit and near-miss fractions add up to one. The vulnerability and readiness are plotted as "below-median" and "above-median", and changes relative to the historical period are classified into "increase" (>0) and "decrease" (<0).

### 3.1.2. Univariate Extremes

Evaluating the model's generalizability to return periods beyond the historical record is inherently challenging due to data limitations. To overcome this, we use the withheld 660-year CESM2 piControl simulation (Figure S3 in Supporting Information S1) mentioned above to establish a long, stationary reference distribution, allowing us to directly estimate multi-centennial return levels well beyond the short 36-year training sample (Figure 4a).

We find that our model improves its ability to simulate and generalize extreme events after incorporating the knowledge of spatial extremal dependence structure. Return-level analysis of annual maxima of spatially averaged temperature shows that DeepX-GAN-CESM2 generates extremes extending beyond the training record while remaining similar to the withheld reference distribution up to multi-centennial scales, while the baseline model exhibits increasing upward bias at return periods exceeding 120 years (Figure 4a). We also find that return levels estimated by a generalized extreme value (GEV) distribution fitted solely to the 36-year short training sample are highly sensitive to sampling variability, leading to huge uncertainties at longer return periods (yellow shading in Figure 4a). This highlights the intrinsic difficulty of reliably estimating "true" long-return-period extremes from limited historical records alone. Across repeated experiments with different training segments, the ensemble spread of return levels generated by DeepX-GAN-CESM2 remains similar to the reference distribution derived from the withheld simulation (Figure S6 in Supporting Information S1), indicating stable extrapolation and low sensitivity to sampling variability—a significant advantage over GEV fitting, which exhibits substantial uncertainty when applied to short observational records (Figures S7–S11 in Supporting Information S1).

#### 3.1.3. Spatial Extremes: Clustering Pattern

Reliable unseen generation further requires preserving the joint structure of spatial extent and intensity of extremes. We therefore examine spatially connected extremes by identifying clusters in which each pixel exceeds its local 95th-percentile threshold. We illustrate how intensity varies with cluster size (Figure 4b). Both DeepX-GAN-CESM2 and the baseline model reproduce the median and range of intensity levels that are similar to those in the training and test data from small to medium cluster sizes. At larger cluster areas, however, the baseline model extends to spatial extent beyond those shown in the test data set (area fraction >0.19). Because the withheld 660-year CESM2 simulation provides a long reference record of physically simulated variability, cluster sizes beyond its range are unlikely to be physically plausible. In contrast, DeepX-GAN-CESM2 does not show such large-area extremes, with its cluster areas remaining within the range supported by the test data set. The range envelopes indicate that DeepX-GAN-CESM2 spans a wider range of intensities than the training data, while remaining bounded by the test data distribution (Figure 4b; blue shading extends beyond yellow dashed lines but remains within the red envelope). This supports that DeepX-GAN-CESM2 preserves the scaling relationship between spatial extent and intensity, without producing unrealistic combinations of large extent and high intensity. These results are consistent across repeated experiments with different training segments (Figures S8–S10 in Supporting Information S1).

While the DeepX-GAN-CESM2 generated clusters are not designed (and therefore expected) to reproduce individual events in the CESM2 simulation exactly, they exhibit comparable spatial organization and coherence, including contiguous regions of elevated intensity and realistic spatial gradients (Figures 4g–4j). For example, we observe warm anomalies over Southern Europe in both DeepX-GAN-CESM2 (Figure 4g) and CESM2 simulations (Figure 4h), possibly linked with anticyclonic circulation and atmospheric blocking (Kautz et al., 2022; Liu et al., 2020). Other examples feature warming clusters over North Africa, West Asia, and India (Figures 4i and 4j), possibly associated with anomalous stationary Rossby waves (Govardhan et al., 2025; Kornhuber et al., 2020).

#### 3.1.4. Spatial Extremes: Dependence Structure

We also evaluate whether generated extremes preserve the dependence structure across space that is critical for compound risk representation. The spatial extremal dependence statistics we compute here are pairwise extreme correlations across pixelwise locations. DeepX-GAN-CESM2 more closely matches both training and reference distributions than the baseline model, as is also reflected by higher coefficients of determination and lower errors (Figures 4c–4f). When evaluating against the test set instead of the training set, the spatial dependence structure is still reasonably reconstructed (Figures 4c–4f, Table S2 and Figures S7–S11 in Supporting Information S1). This indicates that generated extremes retain a realistic spatial dependence structure even when evaluated on independent, withheld data.

In summary, these out-of-sample evaluations against physics-based, long-term CESM2 simulations demonstrate that the generated unseen extremes exhibit realistic tail behavior, spatial coherence, and physically plausible large-scale organization, supporting the model's reliability in extrapolating beyond limited observational records. A possible explanation is that when extremes occur at a particular location, our generative model implicitly leverages information from spatially correlated locations through learned extremal dependencies—often governed by large-scale circulation patterns. This spatial information-sharing mechanism enables the model to capture coherent spatial footprints of clustered extremes without explicitly modeling underlying physical drivers, making it possible for statistically plausible and physically consistent extreme event generation.

### 3.2. Unseen "Direct-Hit" and "Near-Miss" Events

Impacts of unseen extremes on a target location may arise not only from direct local extremes but also from nearby extremes. Our model captures these events by generating spatially structured extreme fields across the full domain. We first quantify the probability of community-wide unseen extremes (Figure 2a), defined as the likelihood of exceeding an extreme threshold anywhere within a community encompassing both the target location and its adjacent areas. We then disentangle how such extremes may specifically affect a target location through two complementary pathways: by directly exceeding the threshold at the location (unseen direct-hit extreme; Figure 2b), or by impacting neighboring areas and indirectly jeopardizing the target location (unseen near-miss extreme; Figure 2c). To enable consistent comparisons across locations, we quantify direct-hit and near-miss extremes by fractions of the community-wide unseen extremes, such that the two fractions sum to one for each target location.

DeepX-GAN-ERA5 is trained on ERA5 reanalysis daily maximum temperature in boreal summer (June-July-August) from 1979 to 2014 to learn the spatial distribution of extremes and to enable the generation of statistically plausible but historically unobserved events. Our analysis reveals pronounced spatial heterogeneity in the probability of community-wide unseen extremes (Figure 5a), with the most prominent and spatially coherent hotspots located in Northwest Africa and the Arabian Peninsula, alongside more scattered hotspots across Southeast Europe and broader regions of Africa and Asia. Elevated community-wide unseen probabilities in these regions suggest that the historical record is less likely to have fully sampled the upper tail of the distribution. As a result, unprecedented events are statistically more likely to emerge, even under present-day climate conditions. Overlapping partially with the hotspots identified in the community-wide unseen probability, Northwest Africa, the Arabian Peninsula, and Southeast Europe also show elevated direct-hit fractions (Figure 5b). This shows that in these regions, unprecedented events are more likely to directly impact local locations.

Although direct-hit and near-miss events are defined locally, their spatial patterns can reflect the organization of extremes at larger scales. Regions with clustered and smoothly varying direct-hit fractions are likely to be more frequently involved in spatially compounding extremes that occur jointly across neighboring locations, leading to a higher likelihood of direct impacts at these nearby grid points. This implies a higher potential for systematic regional impacts (see Discussion for more details). In contrast, fragmented patterns of elevated direct-hit fraction may be associated with isolated local anomalies. Interpreting these spatial structures therefore provides insight into the scale and interconnectedness of extreme risk.

To place these spatial patterns in context, we also compare the results against a simple null model that assumes spatial independence, in which extreme events occur randomly and independently at each grid point. Under this assumption, the probability of both community-wide and direct-hit unseen extremes would be expected to be relatively homogeneous. Instead, we find pronounced large-scale heterogeneity in both metrics. Relative to the null model, fractions of direct-hit extremes are systematically higher and near-miss lower across over 62.7% of the MENA region (Figure 5b). This deviation indicates that the spatial organization of unseen extremes arises from structured dependence rather than random coincidence.

### 3.3. Future Unseen Extremes and Adaptation Gaps

To examine how unseen extremes may evolve under climate change, we analyze future climate scenarios using CMIP6 CMCC-ESM2 model simulations. DeepX-GAN-CMCC-ESM2 is trained separately on climate model outputs from the historical (1979–2014), SSP126 (2065–2100), and SSP585 (2065–2100) scenarios. This allows each generative model to maintain the statistical properties of its corresponding climate simulation and therefore captures the climate-change signal inherent in the underlying climate model (Note S1 in Supporting Information S1). These future projections should be interpreted with caution, as a single climate model cannot capture the full range of inter-model variability and uncertainty inherent in CMIP6 ensembles. However, the global temperature change simulated by CMCC-ESM2 at the end of the 21st century for the different scenarios is in general agreement with CMIP6 multi-model mean values (Lovato et al., 2022) and does not exhibit the high equilibrium climate sensitivity (i.e., the "too hot" problem) found in several other CMIP6 models (Voosen, 2022). This middle-of-the-road realization suggests our projected extremes are unlikely to be artificially inflated by an oversensitive climate response. Conversely, models with higher climate sensitivity would likely produce more intense and frequent heat extremes. Our projections of unseen heat extremes could therefore be viewed as illustrating one plausible trajectory under a moderate model realization.

We find that future climate change could elevate and redistribute unseen extremes (Figure 6). To quantify this spatial redistribution, we define hotspots of community-wide unseen extremes using a simplified expected exceedance probability (see methods for details). We find that the geographic distribution of hotspots shifts under future scenarios (Figure 6). For community-wide unseen extremes, only about 47% (SSP126; e.g., Northwest Africa) and 37% (SSP585; e.g., the Arabian Peninsula) hotspots are consistent with historical simulation, while about 53% (SSP126; e.g., the African Transition Zone) and 63% (SSP585; e.g., Central Africa) new hotspots emerge in future scenarios. In contrast, unseen direct-hit fractions have smaller-scale geographical shifts. Only 38% (SSP126; e.g., Northern Africa) and 35% (SSP585; e.g., Central Africa) of the regions with a direct-hit fraction larger than the spatial independent null model are emerging hotspots. These shifts underscore that even the mitigated scenario induces substantial unseen event redistribution, particularly elevating unseen probabilities in Northern Africa (SSP126) and Central Africa (SSP585), where historically there is a low likelihood of encountering community-wide unseen extremes or unseen direct-hit extremes.

Overlaying unseen extremes with socioeconomic conditions, we identify an inequity in the distribution: less developed, highly vulnerable countries face disproportionately high probabilities of community-wide unseen extremes despite minimal contributions to global emissions. Among the 12 countries that are most vulnerable and least ready (i.e., in the top quartile of vulnerability and bottom quartile of readiness), 42% rank in the top quartile of community-wide unseen extreme probability (Figure 5c). In contrast, only one of the 16 countries (Hungary) in the opposite tail (i.e., bottom quartile of vulnerability and top quartile of readiness) falls within this group of top-quartile unseen probability (Figure 5d). This suggests that countries with lower adaptability are also likely to face elevated chance of unseen extremes, which is further confirmed by statistical analysis showing a robust ($p < 0.1$) negative correlation between community-wide unseen extreme probability and national readiness indicator (Figure S12B in Supporting Information S1). Such disparity could widen adaptation gaps between developing and developed countries, as vulnerable nations facing elevated community-wide unseen extremes may lack the anticipatory capacity or resources to plan for unprecedented events that fall outside historical experience—thereby exacerbating existing inequalities in climate change impacts (Chancel et al., 2023; Diffenbaugh & Burke, 2019). End-of-century projections (2065–2100) suggest that these relationships will likely not diminish under the mitigated SSP126 scenario (Figure S12C in Supporting Information S1) and remain pronounced under SSP585 scenarios (Figures S12E and S12F in Supporting Information S1). On the contrary, unseen direct-hit and near-miss fractions do not exhibit such significant socioeconomic correlations (Figure 5d). Yet, many regions with high vulnerability and low readiness are projected to experience concurrent increases in community-wide unseen extreme probability and direct-hit fraction (Figure 6). In these regions, constraints in preparedness and adaptive capacity may imply that unprecedented extremes are more likely to exceed existing response capabilities.

## 4. Discussion

The primary challenge in the quantification of extreme events lies in their inherent scarcity. This rarity makes it difficult to derive reliable statistics from finite observational records (AghaKouchak et al., 2013; Davison & Gholamrezaee, 2012). Deep generative models offer a promising solution to augmenting data sets and simulating high-impact, low-likelihood events with less computational burden compared to physics-based models. However, reliable AI modeling of extreme conditions remains challenging due to limited representative training data (M. Zhang et al., 2021; Bhatia et al., 2021). Our study demonstrates that incorporating the knowledge of spatial extremal dependence structure into the learning process enables DeepX-GAN to enhance extremal behaviors of synthetic data. Through out-of-sample validation against physics-based climate large-ensemble simulations, we demonstrate the trustworthiness of our model's generated unprecedented extremes in a zero-shot learning context (Figure 4, Figure S3 in Supporting Information S1), enabling risk assessment of low-probability extreme events that may occur without historical precedent. Although it is challenging to guarantee that the generated unseen heat extremes will occur with certainty, we cannot dismiss the possibility of these historically unobserved extremes.

Such information could be useful for stress-testing applications to prepare for the worst-case climate scenarios (Acharya et al., 2023; Qiu et al., 2022).

Integrating the spatial dependence structure of extreme events, our proposed model DeepX-GAN complements traditional statistical methods in modeling unseen extremes. Conventional statistical approaches like extreme value theory are primarily developed for univariate extremes (Coles, 2001) and focus on marginal behavior at individual locations (Thompson et al., 2023; Thompson et al., 2022; L. Zhang et al., 2024). Different from computationally intensive physics-based climate models, DeepX-GAN efficiently generates extensive data ensembles to explore a broad range of possible extremes (Figure 4). By learning the full temperature distribution with emphasis on extremal dependence, DeepX-GAN can generalize to unprecedented extremes. This capability is crucial for assessing potential hazards in regions that have not yet experienced severe heat extremes or have only encountered near-miss events (Fischer et al., 2023; Gessner et al., 2021; Thompson et al., 2023).

Previous studies find that populations who escape severe heat events may develop a false sense of resilience, interpreting short-term survival as evidence of long-term preparedness (Arvai et al., 2006; Dillon & Tinsley, 2008, 2016; Retchless & Ashley, 2022). This "fortune" of temporary avoidance may sow the seed of future susceptibility by deprioritizing systemic adaptations. Therefore, nations with high community-wide unseen extreme probabilities (e.g., Yemen, Algeria, and Morocco, Figure 5c) may experience heightened impacts if infrastructure standards and emergency response systems are designed or stress-tested solely against historical records. This aligns with the adaptation paradox theory (Moser & Ekstrom, 2010; Pelling et al., 2015; Wise et al., 2014), where success in avoiding immediate hazards reduces perceived urgency for long-term resilience investments. Conversely, the "misfortune" of recurrent historical exposure to heat extremes could catalyze proactive infrastructure investments and policy reforms.

Regions exhibiting high fractions of unseen direct-hit extremes, such as Algeria and Saudi Arabia (Figure 5b), are likely to experience simultaneous stress when an unprecedented heat extreme affects the broader community. When extreme heat occurs across neighboring regions at the same time, electricity demand for air conditioning can surge concurrently across interconnected grids, increasing the chance of widespread outages. Water demand may rise simultaneously while supply systems face thermal and evaporative losses. Outdoor labor productivity can decline concurrently, affecting construction, transport, and agriculture. Such synchronized stress makes response and recovery more difficult: medical services and cooling shelters may reach capacity, electricity imports or load transfers become constrained under system-wide peak demand, and public budgets must absorb multiple sectoral losses at once. Insurance claims related to health issues, crop losses, and business interruptions may accumulate rather than being staggered over time. This necessitates planning strategies and early-warning systems that integrate regional climate risks rather than relying on localized conditions alone. In contrast, regions with low direct-hit but high near-miss fractions (e.g., Jordan, India, and Pakistan, Figure 5b) may face fewer interdependent threats from neighboring climate hazards. Yet, a warming climate could increase such interdependency events (Figures 6e–6h), emphasizing the importance of regional coordination in climate monitoring and infrastructure investment, even in locations that have historically avoided direct impacts.

Stakeholder risk perception further shapes responses to unseen extreme signals. Risk-averse populations may interpret near-miss events as indicators of latent vulnerability, prompting proactive adaptation (i.e., vulnerable near-miss framing (Dillon & Tinsley, 2016; Retchless & Ashley, 2022)). Conversely, risk-tolerant groups may perceive such near-misses as evidence of systemic resilience, potentially fostering complacency (i.e., resilient near-miss framing (Dillon & Tinsley, 2016; Retchless & Ashley, 2022)). These divergent perspectives could influence national policymaking: economically advantaged nations with high unseen near-miss fractions (e.g., Austria) may use financial flexibility to stress-test infrastructure, adjust insurance capital buffers, and invest preemptively in heat resilience. In contrast, comparable actions remain challenging for resource-constrained countries facing similar near-miss levels (e.g., Jordan, Liberia), as aggressive preparation measures could exacerbate existing fiscal limitations, potentially widening global inequalities in climate preparedness.

Under future warming scenarios (SSP126 and SSP585), the spatial expansion and redistribution of unseen extreme hotspots will likely introduce new sources of uncertainty, particularly in vulnerable regions. This shift underscores the need for spatially adaptive policies that anticipate emergent hotspots rather than extrapolating historical patterns. Additionally, socioeconomic inequities compound the adverse impacts of climate change on unseen extremes: low-readiness, high-vulnerability nations that face enhanced probabilities of unseen extremes (e.g., Yemen and Central African Republic) will likely suffer more from unprecedented heat extremes despite minimal emissions, exemplifying climate injustice (Diffenbaugh & Burke, 2019; Roberts & Parks, 2007; O'Brien & Leichenko, 2000; J. Long, 2024), where climate hazards intersect with preexisting inequalities in resource access. Addressing this intersection of emerging hazards and climate injustice requires scaling up loss and damage financing (Boyd et al., 2017; James et al., 2014; Mechler & Schinko, 2016; Mechler & Schinko, 2016, 2016; Roe et al., 2023) and prioritizing support for adaptive capacity in the most affected regions.

## Conflict of Interest

The authors declare no conflicts of interest relevant to this study.

## Availability Statement

Historical reanalysis of daily maximum 2-m air temperature is aggregated from ERA5 hourly data set (Hersbach et al., 2023). The data used in the unseen reliability evaluation is obtained from the CESM2 Large Ensemble piControl simulations (Danabasoglu et al., 2020). Climate projections are downloaded from the CMIP6 CMCC-ESM2 model simulations for the historical period (Lovato et al., 2021b) and future scenarios (Lovato et al., 2021a). The ND-GAIN vulnerability and readiness indicators can be downloaded from the Notre Dame Global Adaptation Initiative (University of Notre Dame, 2023).

Python scripts used to produce the results in this paper are available from Zenodo (Liu, 2026a) and GitHub (Liu, 2026b).

**Acknowledgments**

This work was supported by the Singapore Ministry of Education (MOE) Academic Research Fund Tier-1 project (A-8001177-00-00) and Tier-2 project (MOE-T2EP50123-0009). X. H. acknowledges support from the Global Climate Research Prize. The computational work for this study was (fully/partially) performed on the resources of the National Supercomputing Centre, Singapore. We thank Alan D. Ziegler for his valuable comments and advice.

## References

Aalbers, E. E., Lenderink, G., van Meijgaard, E., & van den Hurk, B. J. J. M. (2018). Local-scale changes in mean and heavy precipitation in Western Europe, climate change or internal variability? *Climate Dynamics*, *50*(11), 4745–4766. https://doi.org/10.1007/s00382-017-3901-9

Acharya, V. V., Berner, R., Engle, R., Jung, H., Stroebel, J., Zeng, X., & Zhao, Y. (2023). *Climate stress testing*. (Tech. Rep.). Federal Reserve Bank of New York.

AghaKouchak, A., Sellars, S., & Sorooshian, S. (2013). Methods of tail dependence estimation. In A. AghaKouchak, D. Easterling, K. Hsu, S. Schubert, & S. Sorooshian (Eds.), *Extremes in a changing climate: Detection, analysis and uncertainty* (pp. 163–179). Springer Netherlands. https://doi.org/10.1007/978-94-007-4479-0_6

Arvai, J., Gregory, R., Ohlson, D., Blackwell, B., & Gray, R. (2006). Letdowns, wake-up calls, and constructed preferences: People's responses to fuel and wildfire risks. *Journal of Forestry*, *104*(4), 173–181. https://doi.org/10.1093/jof/104.4.173

Bevacqua, E., Suarez-Gutierrez, L., Jézéquel, A., Lehner, F., Vrac, M., Yiou, P., & Zscheischler, J. (2023). Advancing research on compound weather and climate events via large ensemble model simulations. *Nature Communications*, *14*(1), 2145. https://doi.org/10.1038/s41467-023-37847-5

Bhatia, S., Jain, A., & Hooi, B. (2021). ExGAN: Adversarial generation of extreme samples. *Proceedings of the AAAI Conference on Artificial Intelligence*, *35*(8), 6750–6758. https://doi.org/10.1609/aaai.v35i8.16834

Bishop, C. M., & Bishop, H. (2024). Generative adversarial networks. In C. M. Bishop & H. Bishop (Eds.), *Deep learning: Foundations and concepts* (pp. 533–545). Springer International Publishing. https://doi.org/10.1007/978-3-031-45468-4_17

Boulaguiem, Y., Zscheischler, J., Vignotto, E., Wiel, K. v. d., & Engelke, S. (2022). Modeling and simulating spatial extremes by combining extreme value theory with generative adversarial networks. *Environmental Data Science*, *1*, e5. https://doi.org/10.1017/eds.2022.4

Boyd, E., James, R. A., Jones, R. G., Young, H. R., & Otto, F. E. L. (2017). A typology of loss and damage perspectives. *Nature Climate Change*, *7*(10), 723–729. https://doi.org/10.1038/nclimate3389

Brönnimann, S., Franke, J., Valler, V., Hand, R., Samakinwa, E., Lundstad, E., et al. (2025). Past hydroclimate extremes in Europe driven by Atlantic jet stream and recurrent weather patterns. *Nature Geoscience*, *18*(3), 246–253. https://doi.org/10.1038/s41561-025-01654-y

Cai, F., Liu, C., Gerten, D., Yang, S., Zhang, T., Li, K., & Kurths, J. (2024). Sketching the spatial disparities in heatwave trends by changing atmospheric teleconnections in the Northern Hemisphere. *Nature Communications*, *15*(1), 8012. https://doi.org/10.1038/s41467-024-52254-0

Chancel, L., Bothe, P., & Voituriez, T. (2023). *Climate inequality report 2023*. (Tech. Rep.). World Inequality Lab Study 2023/1.

Cherchi, A., Fogli, P. G., Lovato, T., Peano, D., Iovino, D., Gualdi, S., et al. (2019). Global mean climate and main patterns of variability in the CMCC-CM2 coupled model. *Journal of Advances in Modeling Earth Systems*, *11*(1), 185–209. https://doi.org/10.1029/2018MS001369

Coles, S. (2001). *An introduction to statistical modeling of extreme values*. Springer. https://doi.org/10.1007/978-1-4471-3675-0_8

Danabasoglu, G., Deser, C., Rodgers, K. B., & Timmermann, A. (2020). CESM2 large ensemble [Dataset]. *NSF National Center for Atmospheric Research*. https://doi.org/10.26024/kgmp-c556

Davison, A. C., & Gholamrezaee, M. M. (2012). Geostatistics of extremes. *Proceedings of the Royal Society A: Mathematical, Physical and Engineering Sciences*, *468*(2138), 581–608. https://doi.org/10.1098/rspa.2011.0412

De Marzo, G., Gabrielli, A., Zaccaria, A., & Pietronero, L. (2022). Quantifying the unexpected: A scientific approach to black swans. *Physical Review Research*, *4*(3), 033079. https://doi.org/10.1103/PhysRevResearch.4.033079

Diffenbaugh, N. S., & Burke, M. (2019). Global warming has increased global economic inequality. *Proceedings of the National Academy of Sciences*, *116*(20), 9808–9813. https://doi.org/10.1073/pnas.1816020116

Dillon, R. L., & Tinsley, C. H. (2008). How near-misses influence decision making under risk: A missed opportunity for learning. *Management Science*, *54*(8), 1425–1440. https://doi.org/10.1287/mnsc.1080.0869

Dillon, R. L., & Tinsley, C. H. (2016). Near-miss events, risk messages, and decision making. *Environment Systems and Decisions*, *36*(1), 34–44. https://doi.org/10.1007/s10669-015-9578-x

Dillon, R. L., Tinsley, C. H., & Burns, W. J. (2014). Near-misses and future disaster preparedness. *Risk Analysis*, *34*(10), 1907–1922. https://doi.org/10.1111/risa.12209

Ding, D., Zhang, M., Pan, X., Yang, M., & He, X. (2019). Modeling extreme events in time series prediction. In *Proceedings of the 25th ACM SIGKDD international conference on knowledge discovery and data mining*.
Donat, M. G., Sillmann, J., & Fischer, E. M. (2020). Chapter 3 - Changes in climate extremes in observations and climate model simulations. From the past to the future. In J. Sillmann, S. Sippel, & S. Russo (Eds.), *Climate extremes and their implications for impact and risk assessment* (pp. 31–57). Elsevier. https://doi.org/10.1016/B978-0-12-814895-2.00003-3
Duan, S., McKinnon, K., & Simpson, I. R. (2025). Quantifying the impact of atmospheric circulation and soil preconditioning with large ensembles of simulation under constrained circulation: A case study of the 2021 Pacific Northwest heatwave. *Authorea Preprints*. https://doi.org/10.22541/essoar.174197875.55659663/v1
Fischer, E. M., Beyerle, U., Bloin-Wibe, L., Gessner, C., Humphrey, V., Lehner, F., et al. (2023). Storylines for unprecedented heatwaves based on ensemble boosting. *Nature Communications*, *14*(1), 4643. https://doi.org/10.1038/s41467-023-40112-4
Fischer, E. M., Sippel, S., & Knutti, R. (2021). Increasing probability of record-shattering climate extremes. *Nature Climate Change*, *11*(8), 689–695. https://doi.org/10.1038/s41558-021-01092-9
Fouillet, A., Rey, G., Wagner, V., Laaidi, K., Empereur-Bissonnet, P., Le Tertre, A., et al. (2008). Has the impact of heat waves on mortality changed in France since the European heat wave of summer 2003? A study of the 2006 heat wave. *International Journal of Epidemiology*, *37*(2), 309–317. https://doi.org/10.1093/ije/dym253
Gessner, C., Fischer, E. M., Beyerle, U., & Knutti, R. (2021). Very rare heat extremes: Quantifying and understanding using ensemble reinitialization. *Journal of Climate*, *34*(16), 6619–6634. https://doi.org/10.1175/JCLI-D-20-0916.1
Gessner, C., Fischer, E. M., Beyerle, U., & Knutti, R. (2023). Developing low-likelihood climate storylines for extreme precipitation over central Europe. *Earth's Future*, *11*(9), e2023EF003628. https://doi.org/10.1029/2023EF003628
Goodfellow, I. J., Pouget-Abadie, J., Mirza, M., Xu, B., Warde-Farley, D., Ozair, S., et al. (2014). *Generative adversarial networks*. MIT Press.
Govardhan, D., Pathak, R., Ashok, K., Asiri, M. I., Zamreeq, A., & Hoteit, I. (2025). Midlatitude circulations linked to seasonal extreme precipitation and extreme temperature events in the Arabian Peninsula. *Climate Dynamics*, *63*(2), 117. https://doi.org/10.1007/s00382-025-07592-0
Hamill, T. M. (2001). Interpretation of rank histograms for verifying ensemble forecasts. *Monthly Weather Review*.
He, X., Pan, M., Wei, Z., Wood, E. F., & Sheffield, J. (2020). A global drought and flood catalogue from 1950 to 2016. *Bulletin of the American Meteorological Society*, *101*(5), E508–E535. https://doi.org/10.1175/BAMS-D-18-0269.1
Hersbach, H., Bell, B., Berrisford, P., Biavati, G., Horányi, A., Muñoz Sabater, J., et al. (2023). ERA5 hourly data on single levels from 1940 to present [Dataset]. *Copernicus Climate Change Service (C3S) Climate Data Store (CDS)*. https://doi.org/10.24381/cds.adbb2d47
Hess, P., & Boers, N. (2022). Deep learning for improving numerical weather prediction of heavy rainfall. *Journal of Advances in Modeling Earth Systems*, *14*(3), e2021MS002765. https://doi.org/10.1029/2021MS002765
Huster, T., Cohen, J. E. J., Lin, Z., Chan, K., Kamhoua, C., Leslie, N., et al. (2021). *Pareto GAN: Extending the representational power of GANs to heavy-tailed distributions*. arXiv e-prints. arXiv:2101.09113.
IPCC. (2022). *Climate change 2022: Impacts, adaptation and vulnerability*. Cambridge University Press. https://doi.org/10.1017/9781009325844
James, R., Otto, F., Parker, H., Boyd, E., Cornforth, R., Mitchell, D., & Allen, M. (2014). Characterizing loss and damage from climate change. *Nature Climate Change*, *4*(11), 938–939. https://doi.org/10.1038/nclimate2411
Jiang, Q., Wang, H.-M., Long, B., Wang, Y., Li, S., Qiu, J., et al. (2026). Complex networks reveal climate models' capability in simulating global synchronized extreme precipitation. *Geophysical Research Letters*, *53*(2), e2025GL118219. https://doi.org/10.1029/2025GL118219
Kautz, L.-A., Martius, O., Pfahl, S., Pinto, J. G., Ramos, A. M., Sousa, P. M., & Woollings, T. (2022). Atmospheric blocking and weather extremes over the Euro-Atlantic sector – A review. *Weather and Climate Dynamics*, *3*(1), 305–336. https://doi.org/10.5194/wcd-3-305-2022
Kelder, T., Müller, M., Slater, L. J., Marjoribanks, T. I., Wilby, R. L., Prudhomme, C., et al. (2020). Using unseen trends to detect decadal changes in 100-year precipitation extremes. *npj Climate and Atmospheric Science*, *3*(1), 1–13. https://doi.org/10.1038/s41612-020-00149-4
Kelder, T., Wanders, N., Van Der Wiel, K., Marjoribanks, T. I., Slater, L. J., Wilby, R. L., & Prudhomme, C. (2022). Interpreting extreme climate impacts from large ensemble simulations—Are they unseen or unrealistic? *Environmental Research Letters*, *17*(4), 044052. https://doi.org/10.1088/1748-9326/ac5cf4
Klemmer, K., Xu, T., Acciaio, B., & Neill, D. B. (2022). SPATE-GAN: Improved generative modeling of dynamic spatio-temporal patterns with an autoregressive embedding loss. *Proceedings of the AAAI Conference on Artificial Intelligence*, *36*(4), 4523–4531. https://doi.org/10.1609/aaai.v36i4.20375
Kochkov, D., Yuval, J., Langmore, I., Norgaard, P., Smith, J., Mooers, G., et al. (2024). Neural general circulation models for weather and climate. *Nature*, *632*(8027), 1060–1066. https://doi.org/10.1038/s41586-024-07744-y
Kornhuber, K., Coumou, D., Vogel, E., Lesk, C., Donges, J. F., Lehmann, J., & Horton, R. M. (2020). Amplified Rossby waves enhance risk of concurrent heatwaves in major breadbasket regions. *Nature Climate Change*, *10*(1), 48–53. https://doi.org/10.1038/s41558-019-0637-z
Larochelle, H., Erhan, D., & Bengio, Y. (2008). Zero-data learning of new tasks. *Proceedings of the Twenty-Third AAAI Conference on Artificial Intelligence*.
Leduc, M., Mailhot, A., Frigon, A., Martel, J.-L., Ludwig, R., Brietzke, G. B., et al. (2019). The ClimEx project: A 50-Member ensemble of climate change projections at 12-km resolution over Europe and Northeastern North America with the Canadian regional climate model (CRCM5). *Journal of Applied Meteorology and Climatology*, *58*(4), 663–693. https://doi.org/10.1175/JAMC-D-18-0021.1
Lehner, F. (2024). Climate model large ensembles as test beds for applied compound event research. *iScience*, *27*(11), 111113. https://doi.org/10.1016/j.isci.2024.111113
Liu, X. (2026a). Archived repository for codes of the paper "capturing unseen spatial heat extremes through dependence-aware generative modeling [Software]. *Zenodo*. https://doi.org/10.5281/zenodo.21511472
Liu, X. (2026b). DeepX-GAN GitHub repository [Software]. *GitHub*. Retrieved from https://github.com/PREP-NexT/DeepX-GAN
Liu, X., He, B., Guo, L., Huang, L., & Chen, D. (2020). Similarities and differences in the mechanisms causing the European summer heatwaves in 2003, 2010, and 2018. *Earth's Future*, *8*(4), e2019EF001386. https://doi.org/10.1029/2019EF001386
Long, B., Zhang, B., & He, X. (2023). Asymmetric response of global drought and pluvial detection to the length of climate epoch. *Journal of Hydrology*, *625*, 130078. https://doi.org/10.1016/j.jhydrol.2023.130078
Long, J. (2024). Reckoning climate apartheid. *Political Geography*, *112*, 103117. https://doi.org/10.1016/j.polgeo.2024.103117
Lovato, T., Peano, D., & Butenschön, M. (2021a). CMCC CMCC-ESM2 model output prepared for CMIP6 CMIP [Dataset]. *Earth System Grid Federation*. https://doi.org/10.22033/ESGF/CMIP6.13164
Lovato, T., Peano, D., & Butenschön, M. (2021b). CMCC CMCC-ESM2 model output prepared for CMIP6 CMIP historical [Dataset]. Earth System Grid Federation. https://doi.org/10.22033/ESGF/CMIP6.13195
Lovato, T., Peano, D., Butenschön, M., Materia, S., Iovino, D., Scoccimarro, E., et al. (2022). CMIP6 simulations with the CMCC Earth system model (CMCC-ESM2). *Journal of Advances in Modeling Earth Systems*, *14*(3), e2021MS002814. https://doi.org/10.1029/2021MS002814

Lovell, R. S. L., Collins, S., Martin, S. H., Pigot, A. L., & Phillimore, A. B. (2023). Space-for-time substitutions in climate change ecology and evolution. *Biological Reviews*, *98*(6), 2243–2270. https://doi.org/10.1111/brv.13004

Maher, N., Milinski, S., & Ludwig, R. (2021). Large ensemble climate model simulations: Introduction, overview, and future prospects for utilising multiple types of large ensemble. *Earth System Dynamics*, *12*(2), 401–418. https://doi.org/10.5194/esd-12-401-2021

Mechler, R., & Schinko, T. (2016). Identifying the policy space for climate loss and damage. *Science*, *354*(6310), 290–292. https://doi.org/10.1126/science.aag2514

Moser, S. C., & Ekstrom, J. A. (2010). A framework to diagnose barriers to climate change adaptation. *Proceedings of the National Academy of Sciences*, *107*(51), 22026–22031. https://doi.org/10.1073/pnas.1007887107

O'Brien, K. L., & Leichenko, R. M. (2000). Double exposure: Assessing the impacts of climate change within the context of economic globalization. *Global Environmental Change*, *10*(3), 221–232. https://doi.org/10.1016/S0959-3780(00)00021-2

Pasch, R. J., Blake, E. S., Cobb, H. D., & Roberts, D. P. (2006). *Tropical cyclone report, Hurricane Wilma* (Tech. Rep.). NOAA/NWS/Tropical Prediction Center/National Hurricane Center.

Pelling, M., O'Brien, K., & Matyas, D. (2015). Adaptation and transformation. *Climatic Change*, *133*(1), 113–127. https://doi.org/10.1007/s10584-014-1303-0

Peng, X., Liu, Z., Ziegler, A. D., Zhang, C., Liu, Z., & He, X. (2025). Beyond the mean: Cold and warm tail temperature trends reveal asymmetric controls on snowpack changes in the Northern Hemisphere. *Water Resources Research*, *61*(11), e2025WR041419. https://doi.org/10.1029/2025WR041419

Price, I., Sanchez-Gonzalez, A., Alet, F., Andersson, T. R., El-Kadi, A., Masters, D., et al. (2025). Probabilistic weather forecasting with machine learning. *Nature*, *637*(8044), 84–90. https://doi.org/10.1038/s41586-024-08252-9

Qiu, J., Liu, B., Yang, F., Wang, X., & He, X. (2022). Quantitative stress test of compound coastal-fluvial floods in China's Pearl River Delta. *Earth's Future*, *10*(5), e2021EF002638. https://doi.org/10.1029/2021EF002638

Rampal, N., Gibson, P. B., Sherwood, S. C., Queen, L. E., Lewis, H., & Abramowitz, G. (2025). Downscaling with AI reveals the large role of internal variability in fine-scale projections of climate extremes. https://doi.org/10.48550/arXiv.2507.06527

Retchless, D., & Ashley, R. (2022). Learning from Hurricane Laura's near miss: Evacuation decision-making under uncertainty (Tech. Rep. No. Report 4).

Roberts, T. J., & Parks, B. C. (2007). Fueling injustice: Globalization, ecologically unequal exchange and climate change. *Globalizations*, *4*(2), 193–210. https://doi.org/10.1080/14747730701345218

Roe, D., Holland, E., Nisi, N., Mitchell, T., & Tasnim, T. (2023). Loss and damage finance should apply to biodiversity loss. *Nature Ecology & Evolution*, *7*(9), 1336–1338. https://doi.org/10.1038/s41559-023-02088-8

Sadegh, M., Ragno, E., & AghaKouchak, A. (2017). Multivariate copula analysis toolbox (mvcat): Describing dependence and underlying uncertainty using a Bayesian framework. *Water Resources Research*, *53*(6), 5166–5183. https://doi.org/10.1002/2016wr020242

Salvadori, G., Michele, C. D., Kottegoda, N. T., & Rosso, R. (2007). *Extremes in nature: An approach using copulas*. Springer. https://doi.org/10.1007/1-4020-4415-1_4

The Economist. (2024). AI firms will soon exhaust most of the internet's data. Retrieved from https://www.economist.com/schools-brief/2024/07/23/ai-firms-will-soon-exhaust-most-of-the-internets-data

Thompson, V., Dunstone, N. J., Scaife, A. A., Smith, D. M., Hardiman, S. C., Ren, H.-L., et al. (2019). Risk and dynamics of unprecedented hot months in South East China. *Climate Dynamics*, *52*(5), 2585–2596. https://doi.org/10.1007/s00382-018-4281-5

Thompson, V., Dunstone, N. J., Scaife, A. A., Smith, D. M., Slingo, J. M., Brown, S., & Belcher, S. E. (2017). High risk of unprecedented UK rainfall in the current climate. *Nature Communications*, *8*(1), 107. https://doi.org/10.1038/s41467-017-00275-3

Thompson, V., Kennedy-Asser, A. T., Vosper, E., Lo, Y. T. E., Huntingford, C., Andrews, O., et al. (2022). The 2021 western North America heat wave among the most extreme events ever recorded globally. *Science Advances*, *8*(18), eabm6860. https://doi.org/10.1126/sciadv.abm6860

Thompson, V., Mitchell, D., Hegerl, G. C., Collins, M., Leach, N. J., & Slingo, J. M. (2023). The most at-risk regions in the world for high-impact heatwaves. *Nature Communications*, *14*(1), 2152. https://doi.org/10.1038/s41467-023-37554-1

Tomczak, J. M. (2022). *Deep generative modeling*. Springer International Publishing. https://doi.org/10.1007/978-3-030-93158-2_4

University of Notre Dame. (2023). Notre Dame global adaptation initiative's (ND-GAIN) country index. https://gain.nd.edu/our-work/country-index/

von Trentini, F., Leduc, M., & Ludwig, R. (2019). Assessing natural variability in RCM signals: Comparison of a multi model EURO-CORDEX ensemble with a 50-member single model large ensemble. *Climate Dynamics*, *53*(3), 1963–1979. https://doi.org/10.1007/s00382-019-04755-8

Voosen, P. (2022). 'Hot' climate models exaggerate Earth impacts. *Science*, *376*(6594), 685. https://doi.org/10.1126/science.adc9453

Wang, H.-M., & He, X. (2025). Spatially synchronized structures of global hydroclimatic extremes. *Nature Water*, *3*(12), 1376–1388. https://doi.org/10.1038/s44221-025-00520-w

Wang, H.-M., Peng, X., & He, X. (2024). Forecasting fierce floods with transferable AI in data-scarce regions. *The Innovation*, *5*(4), 100652. https://doi.org/10.1016/j.xinn.2024.100652

Watt-Meyer, O., Henn, B., McGibbon, J., Clark, S. K., Kwa, A., Perkins, W. A., et al. (2025). ACE2: Accurately learning subseasonal to decadal atmospheric variability and forced responses. *npj Climate and Atmospheric Science*, *8*(1), 205. https://doi.org/10.1038/s41612-025-01090-0

White, R. H., Anderson, S., Booth, J. F., Braich, G., Draeger, C., Fei, C., et al. (2023). The unprecedented Pacific Northwest heatwave of June 2021. *Nature Communications*, *14*(1), 727. https://doi.org/10.1038/s41467-023-36289-3

Wicker, W., Harnik, N., Pyrina, M., & Domeisen, D. I. V. (2024). Heatwave location changes in relation to Rossby wave phase speed. *Geophysical Research Letters*, *51*(14), e2024GL108159. https://doi.org/10.1029/2024GL108159

Wise, R. M., Fazey, I., Stafford Smith, M., Park, S. E., Eakin, H. C., Archer Van Garderen, E. R. M., & Campbell, B. (2014). Reconceptualising adaptation to climate change as part of pathways of change and response. *Global Environmental Change*, *28*, 325–336. https://doi.org/10.1016/j.gloenvcha.2013.12.002

Zhang, L., Risser, M. D., Wehner, M. F., & O'Brien, T. A. (2024). Leveraging extremal dependence to better characterize the 2021 Pacific Northwest heatwave. *Journal of Agricultural, Biological, and Environmental Statistics*, *31*(1), 24–45. https://doi.org/10.1007/s13253-024-00636-8

Zhang, M., Ding, D., Pan, X., & Yang, M. (2021). Enhancing time series predictors with generalized extreme value loss. *IEEE Transactions on Knowledge and Data Engineering*, 1. https://doi.org/10.1109/TKDE.2021.3108831